\documentclass[conference]{IEEEtran}
\usepackage{amsmath}
\usepackage{multirow}
\usepackage{amssymb}
\usepackage{graphicx}
\usepackage{algorithm}
\usepackage{algorithmic}
\usepackage{diagbox}
\usepackage{subcaption}
\usepackage{color}
\usepackage{url}
\begin{document}
\title{SAFE: Spectral Evolution Analysis Feature Extraction for Non-Stationary Time Series Prediction}
\author{\IEEEauthorblockN{Arief Koesdwiady and Fakhri Karray}
\IEEEauthorblockA{Center for Pattern Analysis and Machine Intelligence\\Department of Electrical and Computer Engineering\\
University of Waterloo, Ontario, Canada\\
Email:\{abkoesdw, karray\}@uwaterloo.ca}
}

\maketitle

\begin{abstract}
This paper presents a practical approach for detecting non-stationarity in time series prediction. This method is called SAFE and works by monitoring the evolution of the spectral contents of time series through a distance function. This method is designed to work in combination with state-of-the-art machine learning methods in real time by informing the online predictors to perform necessary adaptation when a non-stationarity presents. We also propose an algorithm to proportionally include some past data in the adaption process to overcome the Catastrophic Forgetting problem. To validate our hypothesis and test the effectiveness of our approach, we present comprehensive experiments in different elements of the approach involving artificial and real-world datasets. The experiments show that the proposed method is able to significantly save computational resources in term of processor or GPU cycles while maintaining high prediction performances.
\end{abstract}
\IEEEpeerreviewmaketitle

\section{Introduction}
\begin{figure}[h]
\centering
\includegraphics[width=0.5\textwidth]{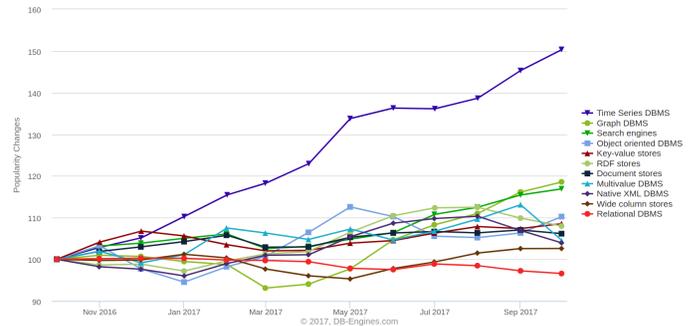}
\caption{The historical trend of the databases popularity.}
\label{timeseries}
\end{figure}

Time series analysis is the study of data that are collected in time order. Commonly, a time series contains a sequence of data that is taken at fixed sampling time. Nowadays, the applications of time-series data are proliferating. For examples, self-driving cars collect data about the environment evolving around them in a continuous manner, and trading algorithms monitor the changing markets to create accurate transaction decisions. According to~\cite{timeseries2017}, time-series databases (TSDBs) have emerged as the fastest growing type of databases for the last 12 months, as can be seen in Figure~\ref{timeseries}.

In general, time series can be categorized into two types: stationary and non-stationary. Roughly speaking, a time series is considered as stationary if its statistical properties remain the same every time. Formally, given a sequence $X_{t1}, \cdots X_{t_k}$ and a sequence $X_{t1+\tau}, \cdots X_{tk+\tau}$ in a time series, if the joint statistical distribution of the first sequence is identical as that of the second sequence for all $\tau$, then the time series is \textit{strictly stationary}~\cite{nason2006stationary}. This means that the moments, e.g., expectations, variance, third-order, and higher, are identical at all times. This definition is extremely strict for real-world applications. Therefore, a weaker definition, namely second-order or weak stationarity, is usually used to analyze time-series for practical applications. Second-order stationary time-series is a time series that has constant mean and variance over time. From this point, a second-order stationary time series is considered as a stationary time series.

The stationarity assumption is especially appealing in time series analysis due to the widely available models, prediction methods, and well-established theory. However, applying this assumption to real-world data, which mostly are non-stationary, might lead to inappropriate forecasts. One of the solutions to handle non-stationarity is to consider non-stationary time series as a collection of piece-wise, or locally, stationary time-series. This means the parameters of the time series are changing but remain constant for a specific period of time. In relation to prediction or forecasting problems, using the piece-wise stationarity assumption, we can employ stationary methods for the prediction and update the predictor when the time-series move to a different stationary state. Therefore it is imperative to continuously check whether the time series is stationary or not.

A vast array of tools for detecting non-stationarity has been introduced by researchers. Most of the detection mechanisms are based on spectral densities~\cite{priestley1969test, adak1998time}, covariance structures comparisons~\cite{andreou2009structural, berkes2009testing}, and, more recently, using locally stationary wavelet model~\cite{cho2012multiscale, korkas2017multiple}. These tests are developed based on a specific choice of segments of the data, which is sometimes delicate and highly subjective. In~\cite{dwivedi2011test}, the test is developed based on the discrete Fourier transform using the entire length of the data, which is undesirable in online settings.

In this work, we are interested in developing a non-stationarity detection method that can be used in real-time and combined with powerful predictors such as state-of-the-art machine learning techniques. In the machine learning community, researchers are more interested in \textit{Concept Drift} detection since most of them are dealing with classification problems~\cite{fdez2007applying,ross2012exponentially, gonccalves2013rcd}. However, in regression problems such as time-series predictions, it is more suitable to consider non-stationarity. Non-stationarity is a more general concept in a sense that time-series without concept drift might contain non-stationarity, e.g., a near-unit-root auto-regressive process. Although the concept is not drifting, i.e., the parameters of the model are static, the time series evolution might contain changing mean and variance. Here, we treat non-stationarity as concept drift and can be used interchangeably.

Generally, there are two ways to detect concept drift: passive and active methods~\cite{ditzler2015learning}. In the passive method, predictors adaptation is done regularly, independent of the occurrence of non-stationarity~\cite{kolter2007dynamic, guajardo2010model, elwell2011incremental}. These methods are quite effective for prediction with gradual changes in the data. However, the main issues are the resource consumption and the potential of overfitting.

On the other hand, the active detection methods monitor the data to detect changes and perform adaptation only when it is needed. These can be done either by monitoring the error of the predictor~\cite{moreira2016concept} or monitoring the features of the data~\cite{cavalcante2016fedd, alippi2011just, liu2013change, alippi2013just}. The main issue of the first approach is that the error might not reflect the non-stationarity of the data and it heavily relies on the prediction accuracy of the predictor, which can be misleading if a poor training process is used to build the predictor. In this work, we are interested in developing an active detection mechanism based on the features of the data.

We propose an online non-stationary detection method based on monitoring the evolution of the spectral contents of a time series. Our main hypothesis is that frequency domain features contain more information than time domain ones. Furthermore, we specifically develop the method to work in combination with state-of-the-art machine learning methods such as Deep Neural Networks (DNN). By combining the power of frequency domain features and the known generalization capability and scalability of DNN in handling real-world data, we hope to achieve high prediction performances.

However, it is known that connectionist models are subjected to a serious problem known as \textit{Catastrophic Forgetting}, i.e., forgetting the previously learned data when learning new data~\cite{french1999catastrophic}. Researchers have been trying to combat this problem by using ensemble learning methods~\cite{rusu2016progressive, polikar2001learn++}, evolutionary computing~\cite{fernando2017pathnet}, and focusing on the regularization aspects of the models~\cite{hashimoto2016joint, kirkpatrick2017overcoming}. These methods are mainly tested on classification problems. In regression problems, more specifically real-world time series problem, it is highly possible that the patterns in the past might not appear again in the future, such as the IBM stock closing price prediction problem, and the future data is highly affected by the past data close to the future only. Therefore, we propose an approach to include some previous data in the past that, which size is variable with respect to the degree of non-stationarity.

Our contribution is summarized as follows:
\begin{itemize}
\item We develop an algorithm to detect non-stationarity based on the evolution of the spectral contents of the time series.
\item We develop an online learning framework that combines the detection method with online machine learning methods efficiently.
\item We propose an algorithm to proportionally include some data in the past to handle Catastrophic Forgetting.
\item We present comprehensive experiments to validate our hypothesis and show the effectiveness of our proposed approach. We performed rigorous experiments to test different distance functions to monitor the evaluation of the spectral contents. We are interested in comparing the frequency domain feature extraction performances with the time-domain feature extraction ones. Finally, we show the superiority of the DNN over several machine learning methods.
\end{itemize}

The rest of the paper is organized as follows. Section~\ref{SAFE} describes the main approach developed in this work, namely SAFE. Section~\ref{predictor} explains the mechanism for embedding predictors with SAFE. Section~\ref{experimental-settings} elaborates the datasets, experimental settings, and performance metrics used to validate our hypothesis and show the effectiveness of our proposed framework. Section~\ref{result} presents the experimental results and discussions. Finally, section~\ref{conclusion} concludes the paper and provides directions for further research.

\section{The SAFE Approach\label{SAFE}}
In this section, the proposed SAFE approach is presented. SAFE is a technique for explicitly detecting non-stationarity in time series. This technique monitors the evolution of the local spectral energy of a time series and computes the discrepancy between the energy at present and previous time instances. The discrepancy then is used to test whether non-stationarity presents or not.

SAFE consists of two main modules: feature extraction, and non-stationarity detection modules. In the first module, Short-Time Fourier Transform (STFT) is applied to extract frequency contents of the time series at each instant time. The results of STFT are frequency values in a complex form. Therefore, the spectral energy of each frequency is computed to simplify our calculations.

The second module uses Simple Moving Average (SMA) and Exponentially Weighted Moving Average (EWMA)~\cite{ross2012exponentially} methods to estimate the long-term and immediate responses of the evolution of the spectral energy through a distance function. In other words, the difference of the spectral energy at every instant time is considered rather than the spectral energy itself.

In an online learning setting, an incoming observation together with its past values are concatenated to form a window in which the STFT is performed. The result of the transformation then compared with the previous window using a distance function. This way, changes can be detected as soon as a new observation arrives.

\subsection{Frequency-Domain Feature Extraction}
In the literature, several time-domain statistical features are used to characterize a time series~\cite{cavalcante2016fedd,alippi2011just}. In this work, a frequency domain approach is presented. There are two main hypotheses in this work:
\begin{itemize}
\item In stationary conditions, the extracted features are expected to be stationary, or at least fluctuating around a stationary value. Therefore, whenever this is not the case, it can be deduced that a non-stationarity presents.
\item More information can be gained in the frequency domain than that can be gained in its counterpart. Therefore, it is expected that the non-stationary detection is more accurate, in terms of true positive and detection delay.
\end{itemize}

\begin{figure}[h]
\centering
\includegraphics[width=0.45\textwidth]{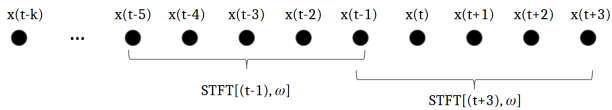}
\caption{An illustration of the STFT process.}
\label{fig:stft}
\end{figure}

In the previous section, it is assumed that non-stationary time-series can be split into chunks of stationary parts. Therefore, a sliding window with a sufficient width can be applied to obtain local stationarity status of a signal. Therefore, it is intuitively suitable to apply STFT to extract the frequency contents of the signal. The discrete STFT can be expressed as

\begin{eqnarray}
\text{STFT}(m,\omega) = \sum^\infty_{k=-\infty}x[k]w[k-m]e^{2\pi j \omega k/L}
\end{eqnarray}
where the $x[k]$ is the time series of interest, $m$, and $\omega$ represent the indicators of time and frequency, respectively; while $L$ is the length of the window function $w$. In this work, a Hamming window function is used. The choice of the sliding window width determines the time-frequency resolution. A wide window provides better frequency and blurred time resolutions while a narrow window works in an inverse way. Once the complex values of each frequency of interest are computed, their spectral energies are then computed. Figure~\ref{fig:stft} illustrates the process of STFT for every sliding window. Take STFT$(t + 3, \omega)$ as an example. When a new observation $x(t+3)$ arrives, the STFT is computed using this values and its several values in the past. It is expected that STFT$(t + 3, \omega)$ and $(t + 2, \omega)$ will be similar if the series is stationary and diverged if it is not the case.

\begin{figure}[h]
\centering
\includegraphics[width=0.5\textwidth]{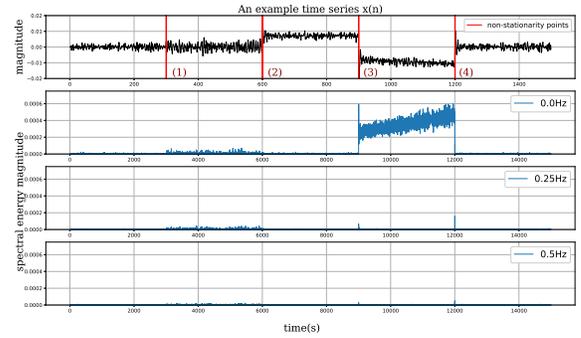}
\caption{An example time series (top) with its spectral energy contents (the rest).}
\label{fig:stft-example}
\end{figure}

To show the effectiveness of the frequency-domain feature extraction on capturing the non-stationarity, a small experiment is conducted and the results are shown in Figure~\ref{fig:stft-example}. In this experiment, four modes of non-stationarity are injected into the time series:
\begin{itemize}
\item The first point, which is denoted by $(1)$ at $t=300$, illustrates a gradual non-stationarity in term of variance.
\item The second point, which is denoted by $(2)$ at $t=600$, shows an abrupt non-stationarity in term of the mean of the series. After this point, the mean is constant, which introduces a bias in the series.
\item The third point, which is denoted by $(3)$ at $t=900$, depicts an abrupt non-stationarity in term of the mean of the series. However, the mean keeps increasing after this point. At this interval, the non-stationarity is in continuous mode.
\item In the last point, the series goes back to the original stationary form.
\end{itemize}

Indeed, there are many modes of non-stationarity that are not included in the experiment. However, these modes are sufficient to illustrate the necessary behavior of non-stationarity for time series prediction. The figure also shows that the spectral energies represent the behavior of the process, which in this case is concentrated in the lower frequency bin. The energy behavior after point (2) looks similar to the one before point(1) although they have different means. This should not be considered as a problem since the important part is the changes at the point of interest, which will be reflected when the distance between points is calculated. It should also be noted that the last point, when the system goes back to the original form, is important to consider since in connectionist modeling the predictor tends to forget the past when a new concept is learned. This creates a problem called \textit{Catastrophic Forgetting}. By keep monitoring the changes, the predictor can be trained to learn the previous concept when necessary.

\subsection{Non-stationarity Detection Module}
The next step after extracting features is to detect whether a non-stationarity occurs or not using the non-stationarity detection module. There are two sub-modules in this module: the distance module, which computes the similarity between to consecutive extracted features, and the non-stationarity test module, which decides whether the observed distances translate to non-stationarity or not. Furthermore, to find the most compatible distance function that can capture the non-stationarity better, three distance functions are tested, namely the Euclidean, absolute Cosine, and absolute Pearson distances. The absolute term is needed since we are interested only in the magnitude of similarity, and not in the direction.


The subsequent step is detecting non-stationarity based on the computed distances, which is developed based on SMA and EWMA. EWMA method estimates the moving mean of a random sequence by giving more importance to recent data. For a set of values of a variable given as $X=\{x_1,x_2, \cdots, x_n\}$, which has mean $\mu_0$, the EWMA estimates the following
\begin{eqnarray}
Z(t) = (1-\lambda)Z(t-1) + \lambda x(t), \;\;\;t>0
\end{eqnarray}
where $Z_0 = \mu_0$. The parameter $\lambda$ weighs the importance of the recent data compared to the older ones.  This is suitable to the proposed feature extraction method, where a new observation is appended to the sliding window and, in the same time, should provide immediate insight that a non-stationarity occurs. In addition, this capability is especially important for detecting gradual non-stationarity. Furthermore, the standard deviation of $Z(t)$ is given as
\begin{align}\label{eq: std}
\sigma_z(t) = \sqrt{\frac{\lambda}{2-\lambda}(1-(1-\lambda)^{2t})\sigma_x}
\end{align}

The output of EWMA represents the immediate estimated mean of the distance while SMA represents the long-term evolution of the mean. Based on this output, a decision about the stationarity has to be made. To do this, an algorithm called SAFE is proposed. This algorithm is illustrated in Algorithm \ref{alg1}. The input of the algorithm is a sequence of data or time series, and initialization  is required for some parameters such as $\lambda$ for the weight of EWMA; $w=0$ for the warning counter; $T$ for the trigger detection threshold; $W$ for the warning detection threshold; and $\gamma$ for the warning duration.

\begin{algorithm}[h]
\caption{SAFE algorithm. \label{alg1}}
\begin{algorithmic}[1]
\renewcommand{\algorithmicrequire}{\textbf{Initialize:}}
\renewcommand{\algorithmicensure}{\textbf{Input:}}
\ENSURE sequence of data
\REQUIRE $\lambda$, $w=0$, $T$, $W$, $\gamma$
\FOR {every instant $t$, a new data $x(t)$ arrives}
\STATE $x_{temp} = [x(t-k), \cdots, x(t-1), x(t)]$
\STATE $f(t) = STFT(x_{temp})$
\STATE $d(t) = dist_{func}(f(t-1), f(t))$
\STATE $Z(t) = (1-\lambda)Z(t-1) + \lambda d(t)$
\STATE compute the SMA of a sufficiently long sliding window of $d(t)$, this is denoted by $\bar \mu(t)$.
\STATE compute $\sigma (t)$ according to Equation \ref{eq: std}.
\IF {$Z(t) \geq \bar \mu (t) + T * \sigma(t)$}
\STATE $ns(t)=1$
\STATE $w=0$
\ELSIF {$Z(t) \geq \bar \mu (t) + W * \sigma(t)$}
\STATE $w = w + 1$
 \IF {$w \geq \gamma$}
\STATE $ns(t)=1$
\STATE $w=0$
 \ENDIF
\ELSIF {$Z(t) \leq \bar \mu (t) + W * \sigma(t)$}
\STATE $w = \max(0, w-1)$
\ENDIF
\ENDFOR
\end{algorithmic}
\end{algorithm}

The first step in the algorithm after initialization is to append new incoming data to the window of previous data to form $x_{temp}$, as depicted in line 2 of the algorithm. Next, STFT is applied to $x_{temp}$, which results in $f(t)$. Subsequently, the distance $d(t$ of $f(t)$ with its previous $f(t-1)$ is computed. In the initial stage of the algorithm, $[x(t-k), \cdots, x(t-1)]$ and $f(t-1)$ are not available. It is safe to assume that $[x(t-k), \cdots, x(t-1)] = [0, 0, \cdots, 0]$ and $f(t-1) = f(t)$ since it is not going to significantly affect the rest of the computation.

\begin{figure}[h]
\centering
\includegraphics[width=0.5\textwidth]{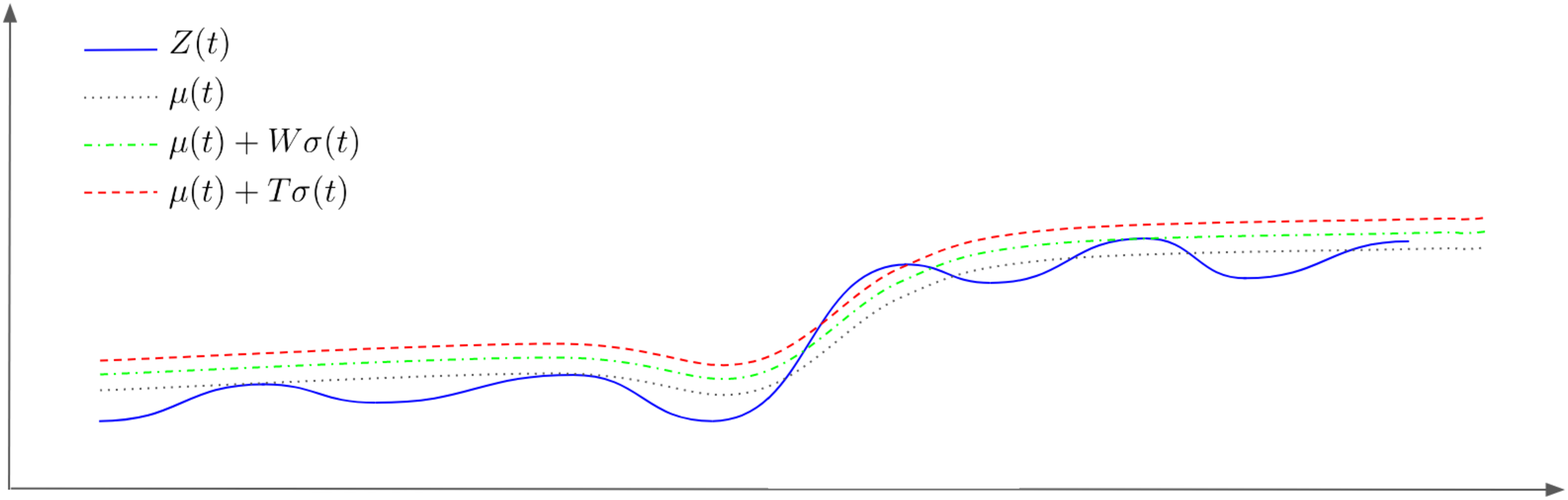}
\caption{Illustration of the SAFE approach.}
\label{fig:error}
\end{figure}

The next step is to apply both EWMA and SMA to $d(t)$, which results in $Z(t)$ and $\bar \mu(t)$. The $\bar \mu(t)$ is necessary since it represents the long-term states of $d(t)$, in particular when a non-stationarity is continuously occurring, as depicted Figure~\ref{fig:stft-example} point 3 to 4. Furthermore, the standard deviation of $Z(t)$ is calculated using Equation \ref{eq: std}. This standard deviation is used together with the control limits $W$ and $T$ as moving thresholds to determine whether $Z(t)$ is still inside a particular stationary mode or not. This idea is illustrated in Figure~\ref{fig:error}. Line 8 and 11 of Algorithm \ref{alg1} impose the moving thresholds to $Z(t)$. If at an instant time $Z(t)$ is greater than $\bar \mu (t) + T * \sigma(t)$, then the non-stationary flag $ns(t)$ is raised. However, when $Z(t)$ is greater than $\bar \mu (t) + W * \sigma(t)$, the algorithm waits until certain duration $\gamma$ to raise the flag. This is also useful when $Z(t)$ is fluctuating insignificantly, which might be due to outliers and/or other unpredictable factors in the data. The flag can be used by predictors to update their parameters when necessary, which is more efficient compared to the blind adaptation scheme.

\section{Embedding SAFE to Online Predictors\label{predictor}}
An online predictor is applied only when a non-stationarity is detected. Initially, a predictor is trained using a presumed stationary dataset in an off-line manner. The parameters obtained from this training are used as initial conditions. In a simulation case, we can select a period of data where the stationary properties hold; while in a real-world case, an initial predictor is trained using all data available at hand.

Once the non-stationarity flag is raised, the next step is to update the parameters of the chosen predictor. The predictor should support online learning since some predictors require training from scratch when new data are available. Some notable online learning algorithms are online passive-aggressive~\cite{crammer2006online}, linear support vector machine (SVM) with stochastic gradient descent (SGD)~\cite{bousquet2008tradeoffs}, and deep neural networks. These learning algorithms are suitable to combine with SAFE. Furthermore, mini-batch SGD also will be more suitable for SAFE since we can include previous data points to form a mini batch and update the predictors accordingly.

It is known that updating these predictors leads to catastrophic forgetting. To avoid this problem we include previous data so that the model also learn the new data using a portion of data from the past. The question is, how many data points should we include in the online adaptation? To answer this question, we introduce the proportional deviation algorithm.

The main idea of this algorithm is to include several numbers of data points proportional to the absolute deviation of $Z(t)$ from $\bar \mu$. Large deviation means the new data is far from the previous stationary point. Therefore, it is intuitively suitable to include more data when the deviation is large and vice versa. The size of the mini batch is computed as follows
\begin{align}
u = \text{round}(\beta |Z(t) - \bar \mu(t)|)
\end{align}
where $u$ is the number of data points included in the past or the size of the mini batch, and $\beta$ is the proportional gain to scale the mini batch. The choice of $\beta$ depends on the applications and affects the speed of the adaptation. The \textit{round} operation rounds the calculation to the nearest integer. This operation is necessary since the size of the mini batch has to be an integer. Algorithm \ref{alg2} is introduced to illustrate this procedure.

\begin{algorithm}[h]
\caption{Embedding SAFE to predictors. \label{alg2}}
\begin{algorithmic}[1]
\renewcommand{\algorithmicrequire}{\textbf{Initialize:}}
\renewcommand{\algorithmicensure}{\textbf{Input:}}
\ENSURE $ns(t)$, $Z(t)$, $\bar\mu(t)$
\REQUIRE $\beta$
\FOR {every instant $t$}
\IF {$ns(t) == 1$}
\STATE $u = \text{round}(\beta \; |Z(t) - \bar \mu(t)|)$
\STATE $x_{train} = [x(t-u), \cdots, x(t-1)]$
\STATE $y_{train}= [y(t-u), \cdots, y(t-1)]$
\STATE $x_{val} = x(t)$
\STATE $y_{val}= y(t)$
\STATE Train/update the chosen predictor using $\{x_{train}, y_{train}\}$ and validate the model using $\{x_{val}, y_{val}\}$.
\ENDIF
\ENDFOR
\end{algorithmic}
\end{algorithm}

In this algorithm, the predictor is trained using $\{x_{train}, y_{train}\}$, and validated using $\{x_{val}, y_{val}\}$. The validation is used to control the training epoch. When the validation error converges, then the training is stopped. This algorithm can be considered as the sub-algorithm of Algorithm \ref{alg1}.

\section{Experimental Settings and Datasets\label{experimental-settings}}
In this section, we describe the datasets and experimental settings that are used to test our hypothesis and to illustrate the effectiveness of the proposed approach. The datasets consist of two types: artificial and real-world data.

Artificial data allow us to perform an effective analysis because the exact locations of non-stationarity are known, which are not the case in real-world data. However, unless we are able to capture all possible non-stationary conditions,  the proposed approaches that are tested using artificial data, might not work well in practice. Therefore, we use real-world data to test the effectiveness our approach. Since the exact locations and behavior of the non-stationarity is not known, we test the real-world data in the prediction stages only. The goal of the test is to see whether the proposed approach is able to correct the prediction that is harmed by non-stationarity.

We propose several experimental settings to test each element in our approach. The settings are defined as follows:
\begin{enumerate}
\item Experiment to find the most suitable distance function for our approach. We investigate three types of distance functions: Euclidean, Pearson, and Cosine distances. This is done using artificial data.
\item Experiment to test whether extracting features in frequency domain leads to better non-stationarity detection results than in time domain. This is done using artificial data.
\item Experiment to investigate which predictor performs well in our approach. This is done using linear and nonlinear artificial data.
\item Experiment to test which domain of feature extraction is better for prediction using real-world data.
\end{enumerate}

At the end of this section, we introduce performance metrics used in each experiment.

\subsection{Datasets}
Here, we describe the datasets used in the experiments.
\subsubsection{Artificial Datasets}
We used two sets of artificial data for experimental settings 1, 2, and 3. The first set contains five time-series data sets that illustrate some behavior of stationary and non-stationary time series. These data sets are inspired, with modifications, by~\cite{cho2012multiscale} and~\cite{korkas2017multiple}, and are given by the following processes


\textbf{(TS-A) Stationary AR(1) process}. This is a stationary process with $\epsilon_t\thicksim\mathcal{N}(0,1)$, and a range of values of $\alpha=\{0.7, 0.4, 0.1, -0.1,-0.4,-0.7\}$.
\begin{align}
x_t = \alpha x_{t-1} + \epsilon_t, \forall 1 \leq t \leq T
\end{align}

\textbf{(TS-B) Piece-wise stationary AR process with obvious changes}. The changes in this process occur at two known locations $t_1=400, t_2=700$.
\begin{align}
x_t = \begin{cases}
0.9 x_{t-1} +\epsilon_t, & \text{if } 1\leq t \leq t_1,\\
1.68 x_{t-1} - 0.81 x_{t-2} + \epsilon_t, & \text{if } t_1 < t \leq t_2,\\
1.32 x_{t-1} - 0.91 x_{t-2} + \epsilon_t, & \text{if } t_2 < t \leq T
\end{cases}
\end{align}

\textbf{(TS-C) Piece-wise stationary AR process with less obvious changes}. The change in this process is less observable and occur at $t_3=600$.
\begin{align}
x_t = \begin{cases}
0.4 x_{t-1} +\epsilon_t, & \text{if } 1\leq t \leq t_3,\\
0.6 x_{t-1}+ \epsilon_t, & \text{if } t_3 < t \leq T
\end{cases}
\end{align}

\textbf{(TS-D) Piecewise stationary near-unit-root process with changing variance}. This process has changes in variance occur at two points $t_4=400, \;t_5 =750$, where $\epsilon_{2t} \sim \mathcal{N}(0, 1.5^2), \epsilon_{3t}\sim \mathcal{N}(0, 3^2)$.
\begin{align}
x_t = \begin{cases}
0.999 x_{t-1} +\epsilon_t, \;\;& \text{if } 1\leq t \leq t_4\\
0.999 x_{t-1}+ \epsilon_{2t}, \;\;& \text{if } t_4 < t \leq t_5\\
0.999 x_{t-1}+ \epsilon_{3t}, \;\;& \text{if } t_5 < t \leq T
\end{cases}
\end{align}

\textbf{(TS-E) Piecewise stationary ARMA(1, 1) process}. This process has three points of changes $t_6=250, t_7=500, t_8=750$.
\begin{align}
x_t = \begin{cases}
0.9 x_{t-1} +\epsilon_t - 0.5 \epsilon_{t-1}, & \text{if } 1\leq t \leq t_6\\
0.3 x_{t-1} + \epsilon_t, & \text{if } t_6 < t \leq t_7\\
0.7 x_{t-1}+ \epsilon_t + 0.6 \epsilon_{t-1}, & \text{if } t_7 < t \leq t_8\\
0.4 x_{t-1} + \epsilon_t - 0.1 \epsilon_{t-1}, & \text{if } t_8 < t \leq T
\end{cases}
\end{align}

The second set of artificial data is inspired by~\cite{cavalcante2016fedd}. This set consists of two linear time series and two non-linear time series; both linear and non-linear time series are known to have parameter changes in some specified points. The linear datasets have similar structure, which is given by AR(p) process. Time series linear-1 and linear-2 are respectively given by AR(4) and AR(5) processes. AR(p) process is given as follows:
\begin{align}
x_t = \alpha_1 x_{t-1} + \alpha_2 x_{t-2} + \cdots + \alpha_p x_{t-p} + e_t
\end{align}
where $e_t\thicksim \mathcal{N}(0, \sigma^2)$.

The nonlinear time series are constructed using the following processes:
\begin{align}\label{eq: nl1}
x_t =& \left[\alpha_1 x_{t-1} + \alpha_2 x_{t-2} + \alpha_3 x_{t-3} + \alpha_4 x_{t-4}\right]\nonumber\\
& * (1 - exp(-10x_{t-1})^{-1} + \epsilon_t)\\
x_t =& \alpha_1 x_{t-1} + \alpha_2 x_{t-2} + \left[\alpha_3 x_{t-1} + \alpha_4 x_{t-2}\right]\nonumber\\
& * (1 - exp(-10x_{t-1})^{-1} + \epsilon_t)\label{eq: nl2}
\end{align}
where Equation \ref{eq: nl1} and Equation \ref{eq: nl2} represent nonlinear-1 and nonlinear-2 time series, respectively. Tabel \ref{parametersnl} presents the parameters used in all the linear and nonlinear time series. Column \textit{Point} provides points where the parameters are implemented.
\begin{table}[h]
\caption{Linear and Nonlinear Time Series Parameters.\label{parametersnl}}
\begin{tabular}{l|c|c|c}
\hline
\hline
Time Series & Point & $\alpha$ & $\sigma^2$\\
\hline
\multirow{4}{*}{Linear-1} & 1-3000 & $\{0.9, -0.2, 0.8, -0.5\}$ & 0.5\\
\cline{2-4}
& 3001-6000 & $\{-0.3, 1.4, 0.4, -0.5\}$ & 1.5\\
\cline{2-4}
& 6001-9000 & $\{1.5, -0.4, -0.3, 0.2\}$ & 2.5\\
\cline{2-4}
& 9001-12000 & $\{-0.1, 1.4, 0.4, -0.7\}$ & 3.5\\
\hline
\multirow{4}{*}{Linear-2} & 1-3000 & $\{1.1, -0.6, 0.8, -0.5, -0.1, 0.3\}$ & 0.5\\
\cline{2-4}
& 3001-6000 & $\{-0.1, 1.2, 0.4, 0.3, -0.2, -0.6\}$ & 1.5\\
\cline{2-4}
& 6001-9000 & $\{1.2, -0.4, -0.3, 0.7, -0.6, 0.4\}$ & 2.5\\
\cline{2-4}
& 9001-12000 & $\{-0.1, 1.1, 0.5, 0.2, -0.2, -0.5\}$ & 3.5\\
\hline
\multirow{4}{*}{Nonlinear-1} & 1-3000 & $\{0.9, -0.2, 0.8, -0.5\}$ & 0.5\\
\cline{2-4}
& 3001-6000 & $\{-0.3, 1.4, 0.4, -0.5\}$ & 1.5\\
\cline{2-4}
& 6001-9000 & $\{1.5, -0.4, -0.3, 0.2\}$ & 2.5\\
\cline{2-4}
& 9001-12000 & $\{-0.1, 1.4, 0.4, -0.7\}$ & 3.5\\
\hline
\multirow{4}{*}{Nonlinear-2} & 1-3000 & $\{-0.5, 0.8, -0.2, 0.9\}$ & 0.5\\
\cline{2-4}
& 3001-6000 & $\{-0.5, 0.4, 1.4, -0.3\}$ & 1.5\\
\cline{2-4}
& 6001-9000 & $\{0.2, -0.3, -0.4, 1.5\}$ & 2.5\\
\cline{2-4}
& 9001-12000 & $\{-0.7, 0.4, 1.4, -0.1\}$ & 3.5\\
\hline
\hline
\end{tabular}
\end{table}
\subsubsection{Real-World Datasets}
We use two real-world datasets to test our proposed approach. The first one is IBM stocks closing price, and the second one is traffic flow of freeways in California.

The IBM dataset was gathered from Yahoo Finance historical data. We collected the stock daily closing price from January 8\textsuperscript{th} 1962 to September 5\textsuperscript{th} 2017, which is around 14000 data points were collected.

The traffic flow dataset was obtained from the Caltrans Performance Measurements Systems (PeMS)~\cite{PeMS}. The traffic flow was sampled every 30 seconds using inductive-loop deployed throughout the freeways. These data were aggregated into 5-min duration by PeMS. Furthermore, the traffic data are aggregated further into 15-min duration based on the recommendation of Highway Capacity Manual 2010~\cite{manual2010volumes}. We collected the traffic flow data of a freeway from January 1\textsuperscript{st} 2011 to December 31\textsuperscript{st} 2012.

\subsection{Experimental Settings}
We apply the datasets to four different experimental settings. In the first setting, we use TS-B to TS-C to find which distance function gives us the best performance on clearly and not so clearly observable changes. In this part, we also illustrate the evolution of each distance function in relation to the time series. Each dataset is tested over 100 trials to account for the randomness introduced by the noise. To gain insight on the performance, we measure the total number of detection, false alarm, hit rate, missed detection, specificity, detection delay, and execution time.

In the second setting, we validate our hypothesis that by extracting the features in frequency domain we can get more information that in the time domain. We use TS-A to measure the false alarm and TS-B to TS-E to measure the total number of detection, false alarm, hit rate, missed detection, specificity, detection delay, and execution time. In the experiment, each dataset is tested over 100 trials.

For the third setting, we use linear (Linear-1, Linear-2) and nonlinear (Nonlinear-1, Nonlinear-2) time series with changes to test the prediction performance when SAFE is embedded into three different predictors namely Passive-Aggressive Regressors, Kernel Linear-SVM, and Deep Neural Networks. In the Kernel Linear-SVM, the original features are mapped into a randomized low-dimensional feature space, which is then trained using linear SVM~\cite{rahimi2008random}. In this experiment, we measure the mean-squared error (MSE), execution time, and percentage update required. We run 20 simulations to account for the randomness.

Finally, we use real-world datasets, namely IBM and Traffic Flow, to measure the performance of deep neural networks which are combined with time-domain feature extraction and SAFE. In this experiment, we observe the MSE, execution time, and percentage update required. In addition, we illustrate all the experimental settings performance with graphs showing their responses over time.
\subsection{Evaluation metrics}
In this section, we describe the performance metrics used in the experiments.

\textbf{Total number of detection}. We report the number of points detected in a time-series. The change point is considered true if it within $5\%$ of the sample size. As an example, the number of detection = 1 means there is only 1 change point is detected. This number is then cumulated over 100 trials. We define false positive as FP, true positive as TP, true negative as TN, and false negative as TN.

\textbf{False Alarm Rate}. It measures the number of change points detected that when there are no actual changes occur. Another name for this metric is False Positive Rate, which is given as $FP/(FP + TN)$.

\textbf{Hit Rate}. It measures the proportion of correctly detected changes over the actual number of change points, which is given as $TP/(TP+FN)$.

\textbf{Missed Detection Rate}. It is the number of undetected changes when there are actual change points over the actual number of change points, which is given as $1 - \text{Hit Rate}$.

\textbf{Specificity}. It reflects the number of stationary points classified as stationary points over the actual number of stationary points. This is calculated as $TN/(TN+FP)$

\textbf{Detection Delay}. It measures the distance, in the number of steps, of the detected changes from the actual points.

\textbf{Execution Time}. It measures the average time, in seconds, required to perform an experiment over a specific number of trials.

\textbf{Mean-Squared Error}. This metric measures the similarity between predictions and actual time series. We used two types of MSE: the trajectory of MSE, which is defined as the evolution of the MSE at every instant time, and the overall MSE, which measures the total MSE of the whole test dataset.

\textbf{Percentage of Update}. The number of updates performed over the number of possible updates. As an example, we have a time series with 100 data points, and we start the online update procedure from $t=1$. Let us assume our algorithm updates the predictors 15 times. It means the percentage of the update is 15\%. If we perform blind adaptation scheme, the percentage of the update will be 100\%.

\section{Results and Discussion\label{result}}
In this section, we discuss the experimental results to evaluate the performances of our proposed approach. The results are grouped according to the experimental settings mentioned in the previous section.
\subsection{Distance Functions}
There are various ways to compute a distance between two vectors; the popular ones are Euclidean, Pearson, and Cosine distances. To decide which distance function gives better performance on non-stationary detection, we conducted experiments using TS-B and TS-C datasets. These datasets illustrate the obvious and non-obvious changes that may occur in a time series.

\begin{figure}[h]
\centering
\includegraphics[width=0.5\textwidth]{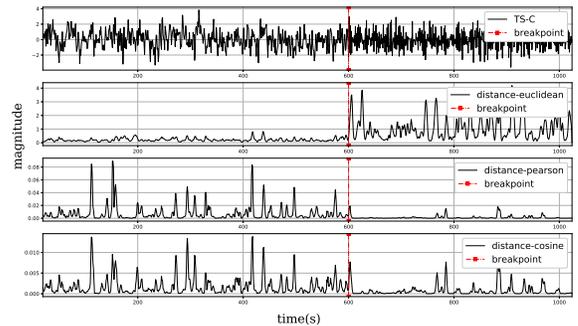}
\caption{Response of distances to TS-C breakpoint}
\label{fig:distance}
\end{figure}

First, it is important to show that these three distance functions can capture changes occur in a time series. Therefore, we present a simulation result showing the responses of the distances to the time series. In this simulation, we applied the SAFE approach to the TS-C dataset with a sliding window of 5 samples. We chose the TS-C dataset because this dataset contains non-obvious changes. Intuitively, when more obvious changes occur, the response of the distance will be more distinguishable. Furthermore, we calculated the distances using the three distance functions mentioned in the previous paragraph. The result of this experiment is shown in Figure~\ref{fig:distance}. This figure illustrates that when a non-obvious change occurs at a breakpoint $t=600$, the three distances respond to the change immediately. It should be noted that the responses of these three distances are different, especially when using the Euclidean distance. Although Pearson and Cosine distances yield similar responses, they are still distinguishable. The response of the Pearson distance is a little bit smoother than that of the Cosine distance.

Next, we investigate the detection performances of the SAFE when it is combined with three distance functions. The size of the sliding window is kept at 5 samples. Other parameters that have to be set is $\lambda, T, W$ and $\gamma$. It is suggested in~\cite{ross2012exponentially} that $\lambda\in [0.1,0.3]$. Since the detection performances highly depend on the thresholds $T, W$, we have to set these thresholds so that three distances have equal performance in one of the evaluation metrics. To achieve this, we set the thresholds such that the three distances have similar false alarm rate on TS-B, which is $\approx 0.05$. From here, we can judge the other performance metrics given the false alarm rate.

To achieve the targeted false alarm rate, we performed several trials. We tried several values of $\lambda$, and we found that $\lambda=0.3$ works best for all the three distances. The warning thresholds for Euclidean, Pearson, and Cosine distances are 2.85, 0.75 and 1.4, respectively, and the trigger thresholds are 3.35, 1,25 and 1.9, respectively. The size of the sliding window for the SMA is 20; this value is able to capture the long-term evolution of the distances. It is expected that these parameters work for different types of non-stationarity. Therefore, these parameters are fixed for all experiments using TS-A to TS-E.

Table \ref{distance} shows the detection performances of the three distances over 100 simulations. We can see that in TS-B and TS-C datasets, Euclidean distance provides superior results in terms of the number of detection, hit rate, missed detection, and execution time. The highest average detection delay on TS-B is produced by Euclidean distance, but this is not the case on TS-C. Overall, Euclidean distance produces better performances than other distances do. Therefore, we chose Euclidean distance as the distance function for the rest of the experiments.

\begin{table*}
\caption {Summary of Non-stationary detection performance using TS-B and TS-C: Euclidean Vs Pearson Vs Cosine distance. Results over 100 simulations.\label{distance}}
\centering
\begin{tabular}{l|c|c|c|c|c|c}
\hline
\hline
\multirow{2}{*}{\backslashbox{Perf.}{dataset}}& \multicolumn{3}{c|}{TS-B} & \multicolumn{3}{c}{TS-C}\\
\cline{2-7}
& euclidean & pearson & cosine & euclidean & pearson & cosine\\
\hline
\# detection: & & & & & \\
\hspace{7mm}0 & 0 & 3 & 0 & 0 & 18 & 4 \\
\hspace{7mm}1 & 4 & 30 & 5 & 100 & 82 & 96\\
\hspace{7mm}2 & 96 & 67 & 95 & - & - & -\\
False Alarm & $0.049$ & 0.050 & 0.050 & $0.051$ & $0.044$ & 0.063 \\
 
Hit rate & $0.98$ & 0.82 & 0.97 & 1.00 & 0.82 & 0.96\\

Missed detection & $0.02$ & 0.18 & 0.03 & 0.00 & 0.12 & 0.04\\

Specificity & 0.95 & 0.95 & 0.95 & 0.95 & 0.96 & 0.95\\

Det. delay (samples) & $11.48\pm 11.62$ & $8.59\pm 8.31$ & $8.23\pm 7.45$ & $6.9\pm 6.48$ & $8.73\pm 9.46$ & $7.83\pm 9.08$\\

Exec. time (s) & $0.19\pm 0.004$ & $0.26\pm 0.008$ & $0.21\pm 0.003$ & $0.19\pm 0.003$ & $0.24\pm 0.007$ & $0.19\pm 0.005$\\
\hline
\hline
\end{tabular}
\end{table*}
\subsection{SAFE Vs Time-domain Feature Extraction}
In the second setting, we compare the performances of SAFE with those of time-domain feature extraction (FE). The time-domain FE is inspired by~\cite{cavalcante2016fedd}. We extracted 4 linear (auto-correlation, variance, skewness coefficient, kurtosis coefficient) and 1 nonlinear (bi-correlation) time-domain features. The detection method after computing the distance is similar to one used by SAFE in Algorithm~\ref{alg1}.

The parameters of the SAFE approach is kept the same as the previous experimental setting, and the parameters of the time-domain FE are set to achieve around 0.05 false alarm rate on TS-B. To achieve this false alarm rate, the forgetting factor $\lambda$ is set to 0.3 while $T$ and $W$ are set to 3 and 3.5, respectively. The sliding window size of the SMA is set to 20. Both SAFE and time-domain FE are experimented using TS-(A-E).

The first experiment on this setting is performed on TS-A. Since TS-A is a stationary time series, it is expected that low false alarm rates are found on both SAFE and time-domain FE. Table \ref{TSA} shows that acceptable false alarm rates are found in both cases. It is evident that using different $\alpha$, SAFE produces lower false alarm rates than time-domain FE does. Indeed, the false alarm rate on stationary time series ideally should be 0. This can be achieved if we set both the thresholds to the higher values. However, setting higher thresholds may lead to poor detection performances. This is up to the designer to decide which performances are important in their applications.

\begin{table}
\caption {Summary of false detection using TS-A. Results over 100 simulations.\label{TSA}}
\centering
\begin{tabular}{c|c|c}
\hline
\hline
\multirow{2}{*}{$\alpha$}& \multicolumn{2}{c}{False alarms}\\
\cline{2-3}
& SAFE & time-domain FE\\
\hline  
0.7 & 0.001 & 0.008\\
  
0.4 & 0.004 & 0.015\\
  
0.1 & 0.011& 0.023\\
 
0.1 & 0.019& 0.030\\
  
0.4 & 0.035 & 0.041\\
 
0.7 & 0.052& 0.053\\
\hline  
Average & 0.020$\pm$0.018 & 0.028$\pm$0.015\\
\hline
\hline
\end{tabular}
\end{table}

The second set of the experiments is done to test the non-stationarity detection performances using TS-(B-E). The results are summarized in Table~\ref{TSs}. This table shows that the overall performance of SAFE on the datasets a better in terms of number detection, false alarm rate, hit rate, missed detection, specificity, detection delay, and execution time. It should be noted that although in some aspects time-domain FE has almost similar performances than SAFE does, time-domain FE has significantly higher average time to execute the experiments. In conclusion, time-domain FE executes the experiments more than 3 times slower than SAFE does.

\begin{table*}[!htbp]
\caption {Summary of Non-stationary detection performance using TS-(B-E): Frequency Vs Time domain. Results over 100 simulations.\label{TSs}}
\centering
\begin{tabular}{l|c|c|c|c}
\hline
\hline
\multirow{2}{*}{\backslashbox{Perf.}{dataset}}& \multicolumn{2}{c|}{TS-B} & \multicolumn{2}{c}{TS-C}\\
\cline{2-5}
& SAFE & time-domain FE & SAFE & time-domain FE\\
\hline
\# detection: & & & & \\
\hspace{7mm}0 & 0 & 0 & 0 & 0\\
\hspace{7mm}1 & 4 & 14 & 100 & 100 \\
\hspace{7mm}2 & 96 & 86 & - & -\\
\hspace{7mm}3 & - & - & - & - \\
False Alarm & $0.049$ & $0.049$ & $0.051$ & $0.050$\\

Hit rate & $0.98$ & $0.96$ & 1.00 & 1.00\\

Missed detection & $0.02$ & $0.04$ & 0.00 & 0.00\\

Specificity & 0.95 & 0.95 & 0.95 & 0.95\\

Det. delay (samples) & $11.48\pm 11.62$ & $12.08\pm 11.50$ & $6.9\pm 6.48$ & $7.17\pm 6.54$\\

Exec. time (s) & $0.19\pm 0.004$ & $0.72\pm 0.013$ & $0.19\pm 0.003$ & $0.73\pm 0.011 $\\
\hline
\hline
\multirow{2}{*}{\backslashbox{Perf.}{dataset}}& \multicolumn{2}{c|}{TS-D} & \multicolumn{2}{c}{TS-E}\\
\cline{2-5}
& SAFE & time-domain FE & SAFE & time-domain FE\\
\hline
\# detection: & & & & \\
\hspace{7mm}0 & 0 & 0 & 0 & 0\\
\hspace{7mm}1 & 8 & 15 & 3 & 1 \\
\hspace{7mm}2 & 92 & 85 & 23 & 30\\
\hspace{7mm}3 & - & - & 74 & 69 \\
False Alarm & 0.039 & 0.041 & 0.042 & 0.047\\

Hit rate & 0.96 & 0.94& 0.94 & 0.88\\

Missed detection & 0.04 & 0.06 & 0.06 & 0.12\\

Specificity & 0.96 & 0.96 & 0.96& 0.95\\

Det. delay (samples) & $11.45\pm 11.66$& $12.38\pm 10.78$ & $15.13\pm 13.61$ & $15.18\pm 14.09$\\

Exec. time (s) & $0.18\pm 0.005$ & $0.72 \pm 0.012$ & $0.19\pm 0.004$ & $0.72\pm 0.010$\\
\hline
\hline
\end{tabular}
\end{table*}

\begin{figure}[!htbp]
\centering
\includegraphics[width=0.5\textwidth]{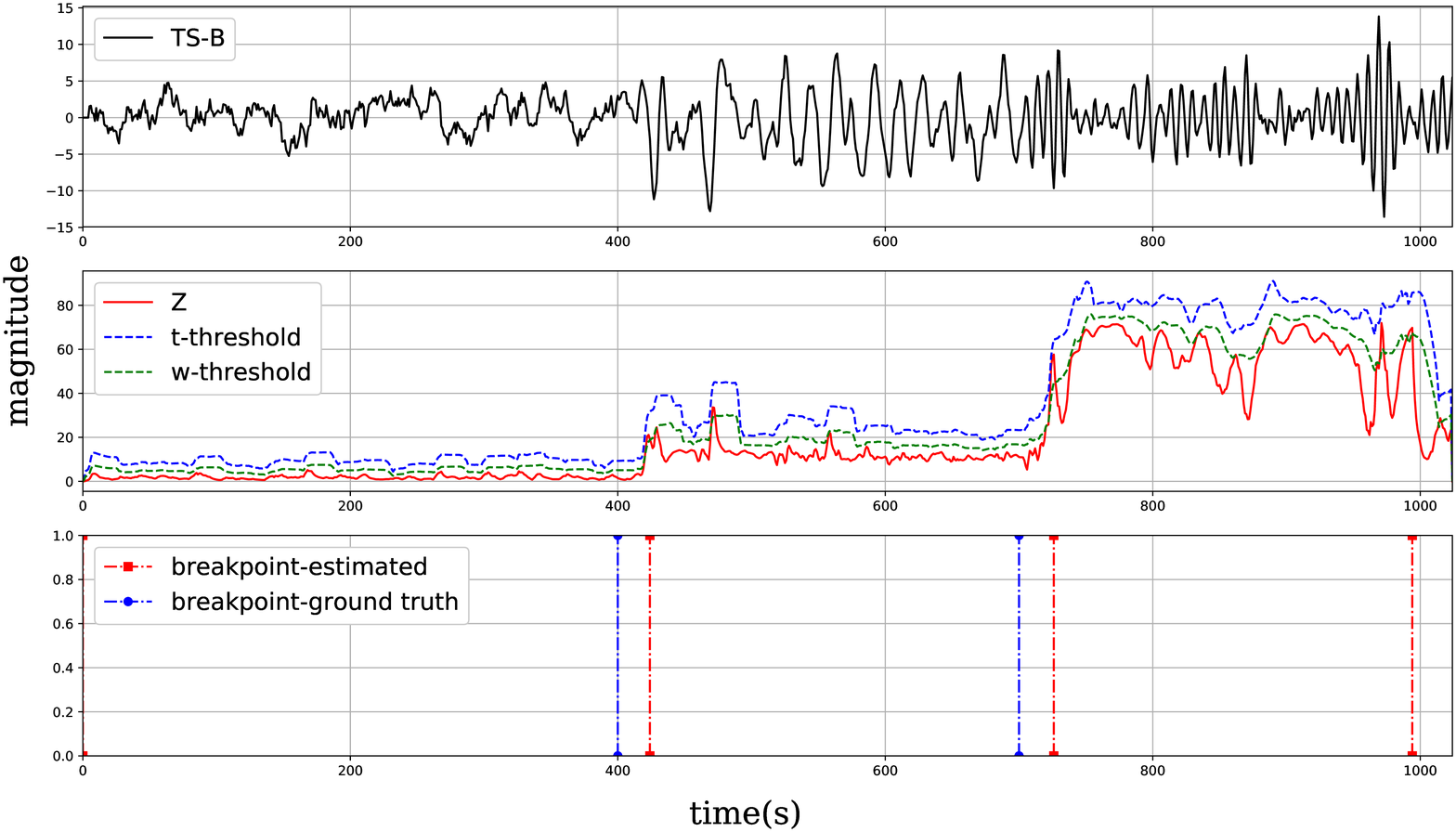}
\caption{Breakpoints detection on TS-B}
\label{fig:B_425}
\end{figure}

\begin{figure}[!htbp]
\centering
\includegraphics[width=0.5\textwidth]{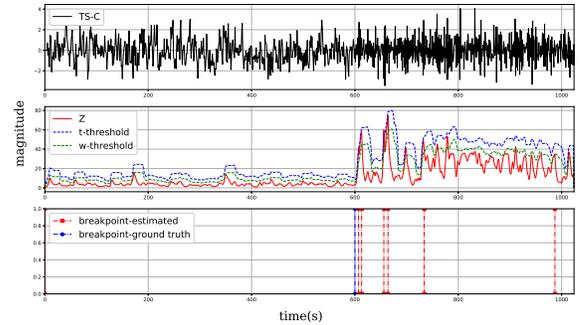}
\caption{Breakpoint detection on TS-C}
\label{fig:C_FA}
\end{figure}

To illustrate the performance of SAFE in more details, we present several graphs showing the predicted change points versus the ground truth. Figure~\ref{fig:B_425} shows the detection on TS-B. From the previous section, we know that TS-B has two change points. However, the figure shows that the proposed algorithm detects three change points, where the third detected point is considered as a false positive. This is acceptable since it can be seen, by inspection from the first row of the graph, that the time series actually looks non-stationary in term of the variance. The second row of the graphs shows the evolution of $Z$ and the warning and trigger thresholds. In Figure~\ref{fig:C_FA}, we can see that similar behavior, where few false alarms present, is also shown in the experiment on TS-C.

\begin{figure}[!htbp]
\centering
\includegraphics[width=0.5\textwidth]{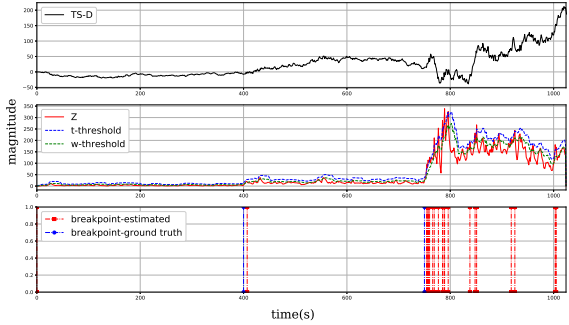}
\caption{Breakpoint detection on TS-D}
\label{fig:D_407}
\end{figure}

Furthermore, different behavior is observable in an experiment using TS-D. Figure~\ref{fig:D_407} shows that the false alarm rate is somewhat higher than the other experiments using different time series. If we look at the behavior of the time series closely, it certainly depicts a non-stationary behavior. This behavior is due to the fact that TS-D is a piece-wise stationary near-unit-root process with changing variance. Near-unit-root process means the process dynamic is close to unstable behavior. Therefore, it is expected that the SAFE approach considers some parts of the time-series as non-stationary, especially in the last part of the series. It is logical to consider that the process contains continuous non-stationarity, rather than just a change in the breakpoint. This way, our chosen online predictor can be continuously updated to adapt to the non-stationary process.

\subsection{Choice of Predictors}
The third setting of the experiments is employed to test which predictor gives better prediction performances. We use Linear-1, Linear-2, Nonlinear-1, and Nonlinear-2 datasets and test them with three different predictors namely Passive-Aggressive Regressors (PAR), Kernel Linear-SVR (KSVR), and Deep Neural Networks (DNN). Before we run the experiments, both parameters in Algorithm~\ref{alg1} and~\ref{alg2} have to be set. We set $\lambda, T, W, \beta$ to be 0.3, 10, 20, and 0.1 respectively.

Furthermore, the parameters of the predictors also need initialization. For PAR, we set the aggressiveness parameter to 0.05. For KSVR, we used epsilon insensitive as the loss function, $L_2$ regularizer with a constant equals to $1e^{-3}$. Finally, for the DNN, we set the number of hidden layers to 2, where each hidden layer contains 200 neurons, and use Rectified Linear Unit (ReLU) as the activation function. To avoid over-fitting, we implemented a drop out regularizer with rate equals to 0.1. All of the parameters were tuned using a similar off-line training scheme. We trained the predictors using the first 2000 samples, and validated using the next 1000 samples. We did not implement K-fold cross-validation since, in time series prediction, it is not suitable to train using data that come after the validation data. Basically, we trained the predictors until the validation errors stop decreasing. The trained predictors are then used as initial models that will be updated when they are triggered by the SAFE algorithm.

Finally, the rest of the data are used to test the predictors. The experimental results of these experiments are summarized in Table~\ref{tab:1a}. The results show that the smallest prediction errors on all datasets are achieved using DNN as the predictor. However, the average execution time of this predictor is also shown to be the highest among all. In some applications, it might be crucial to have slow execution time. Accordingly, if we concern more about execution time than prediction errors, then DNN might not be the best choice as a predictor. However, in an application such as traffic flow prediction, where the sampling time is 15 minutes, DNN is suitable as a predictor since it produces significantly lower prediction errors than the rest of the predictors do.

\begin{figure}[!htbp]
\centering
\includegraphics[width=0.5\textwidth]{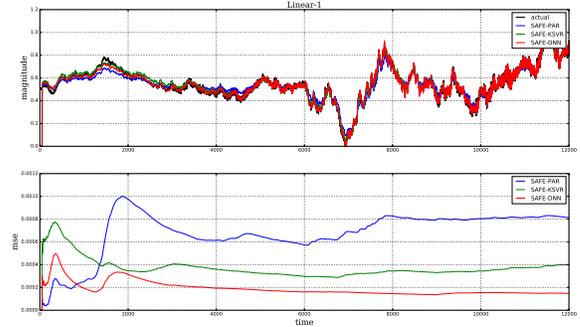}
\caption{Prediction performance on Linear-1 dataset.}
\label{fig:l1}
\end{figure}

\begin{figure}[!htbp]
\centering
\includegraphics[width=0.5\textwidth]{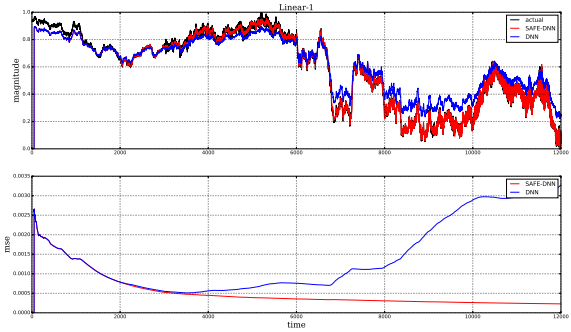}
\caption{Comparison between SAFE-DNN and the baseline predictor on Linear-1 dataset.}
\label{fig:l1ns}
\end{figure}

Figure~\ref{fig:l1} depicts the prediction performances on Linear-1 dataset. The first graph shows that all of the predictors are able to follow the ground truth closely. However, if we look at the errors on the second graph, the DNN error is fairly lower than those of other predictors although the off-line error of the DNN is higher than that of PAR. The next step is to compare the performance of the DNN with the baseline predictor, which is the predictor that is not updated. It can be seen in Figure~\ref{fig:l1ns} that while the error of the DNN stays constant, the error of the baseline predictor is drifting to a larger value. We can conclude that updating the SAFE approach is reliable for non-stationary time series prediction.

\begin{figure}[!htbp]
\centering
\includegraphics[width=0.5\textwidth]{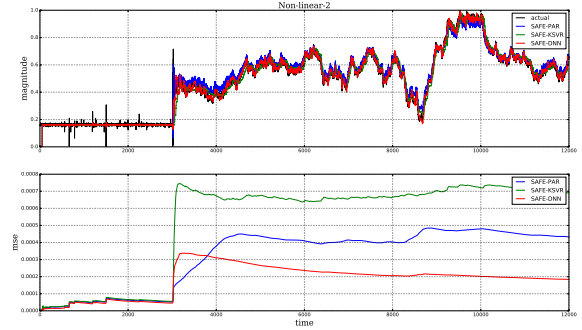}
\caption{Prediction performance on Nonlinear-2 dataset.}
\label{fig:nl2}
\end{figure}

\begin{figure}[!htbp]
\centering
\includegraphics[width=0.5\textwidth]{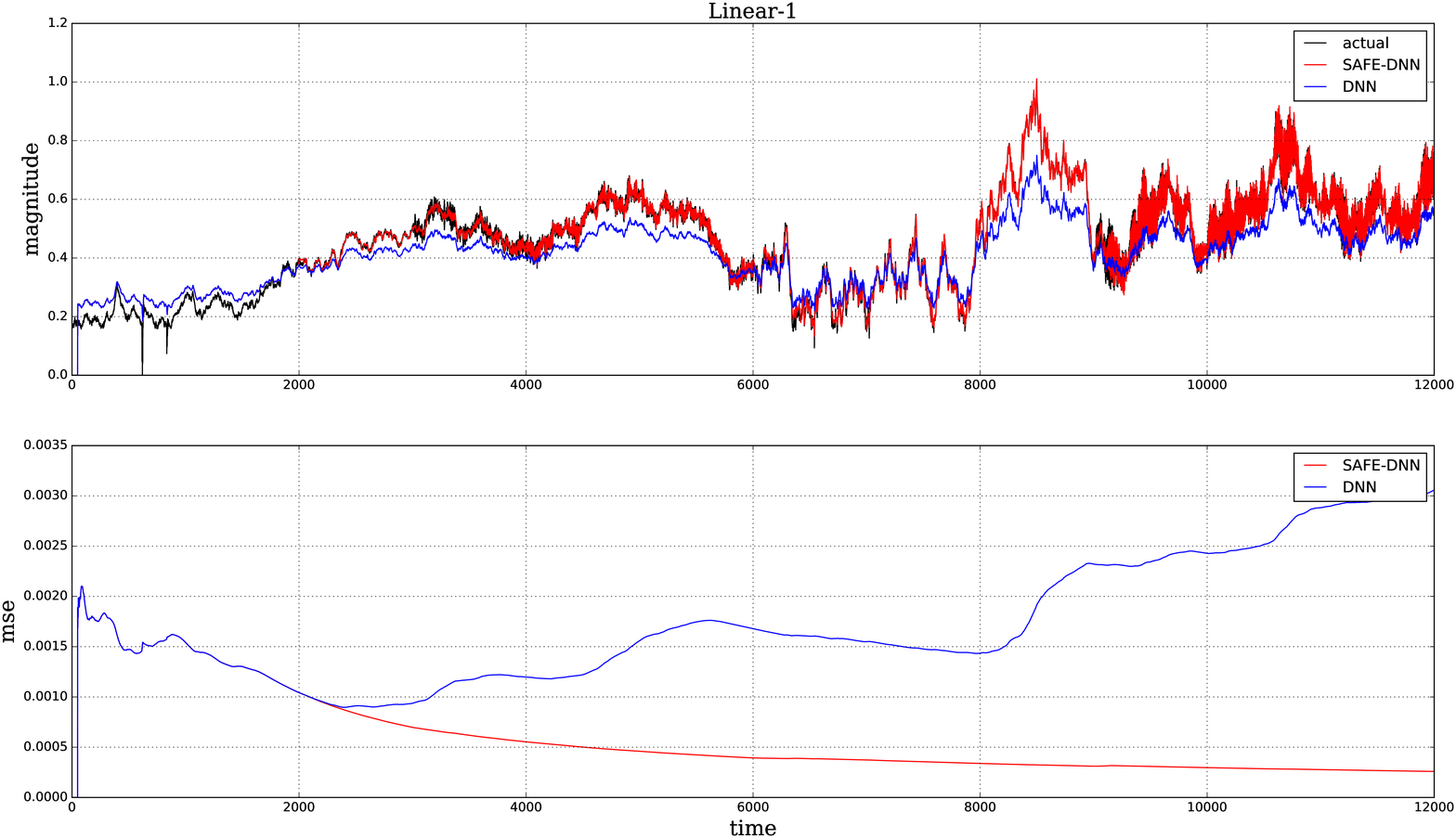}
\caption{Comparison between SAFE-DNN and the baseline predictor Nonlinear-1 dataset.}
\label{fig:nl1ns}
\end{figure}

Lastly, Figure~\ref{fig:nl2} and~\ref{fig:nl1ns} shows that similar performances are shown in the nonlinear datasets. Although the predictor errors start around similar values, the DNN performs better in the long run. In addition, all of these prediction performances are achieved by updating the predictors when necessary only. This is reflected by the percentages of the update for all datasets that are less than 35\%.
\subsection{Real-world Data Experiments}

\begin{table*}[!htbp]
\caption {Summary of Non-stationary time-series prediction performance.}
\begin{subtable}{\linewidth}\caption{Artificial datasets (Results over 20 simulations).}\label{tab:1a}
\centering{\begin{tabular}{l|c|c|c|c|c|c|c}
\hline
\hline
\multirow{2}{*}{Dataset}& \multicolumn{2}{c|}{SAFE-PAR} & \multicolumn{2}{c|}{SAFE-KSVR} & \multicolumn{2}{c|}{SAFE-DNN} & \multirow{2}{*}{\% update}\\
\cline{2-7}
& mse ($\times 10^{-3}$) & ex. time (s) & mse ($\times 10^{-3}$) & ex. time (s) & mse ($\times 10^{-3}$) & ex. time (s) \\
\hline
Linear-1 & $3.70\pm 1.10$ & $3.15\pm 0.22$ & $2.40\pm 1.50$ & $5.30\pm0.92$ & $\mathbf{1.45\pm 1.10}$ & $35.65 \pm 8.59$ & $17.72\pm3.23$\\
Linear-2 & $6.22\pm1.73$ & $3.44\pm0.27$ & $3.20\pm1.42$ & $9.25\pm3.28$ & $\mathbf{2.17\pm1.39}$ & $72.63\pm 24.70$ & $31.03\pm4.07$\\

Non-linear-1 & $4.12\pm 2.42$ & $2.83\pm 0.20$& $1.76\pm1.10$& $5.07\pm 1.38$& $\mathbf{0.90\pm0.91}$& $42.33\pm 12.69$ & $20.39\pm 4.25$\\

Non-linear-2 & $4.20\pm0.92$ & $2.64\pm 0.07$ & $7.15\pm 9.90$ & $3.55\pm0.24$ & $\mathbf{2.17\pm1.00}$ & $26.56\pm5.56$ & $6.56\pm1.68$\\
\hline
\end{tabular}}
\end{subtable}
\begin{subtable}{\linewidth}
\vspace{3mm}
\centering
\caption{Real-world datasets}\label{tab:1b}
{\begin{tabular}{l|c|c|c|c|c|c|c}
\hline
\hline
\multirow{2}{*}{Dataset}& \multicolumn{3}{c|}{Time-DNN} & \multicolumn{3}{c|}{SAFE-DNN} & DNN\\
\cline{2-8}
& mse ($\times 10^{-3}$) & ex. time (s) & \% update & mse ($\times 10^{-3}$) & ex. time (s) & \% update & mse ($\times 10^{-3}$)\\
\hline
\multirow{2}{*}{IBM} & $1.60$ & $28.14$ & $2.7$ & $0.31$ & $17.80$ & $2.7$ & $13.3$\\
& $0.36$ & $46.19$ & $20$ & $0.11$ & $36.23$ & $20$ & $13.3$ \\
\hline
\multirow{2}{*}{Traffic Flow} & $2.24$ & $482.93$ & $5$ & $2.07$ & $434.85$ & $5$ & $4.06$\\
        & $2.04$ & $1518.53$ & $15$ & $\mathbf{1.84}$ & $1420.29$ & $15$ & $4.06$\\
\hline
\hline
\end{tabular}}
\end{subtable}
\end{table*}

The objective of the last set of experiments is to test the prediction performance of DNN on two real-world datasets under SAFE and time-domain FE. The first dataset is IBM dataset. We use the data from January 8\textsuperscript{th} 1962 to January 7\textsuperscript{th} 1977 for training; January 8\textsuperscript{th} 1977 to January 7\textsuperscript{th} 1982 for validation; and January 8\textsuperscript{th} 1982 to September 5\textsuperscript{th} 2017 for testing. The second dataset is traffic flow dataset of San Fransisco Bay Area, District 4, California. We use the data from January 1\textsuperscript{st} 2011 to August 31\textsuperscript{st} 2011 for training; September 1\textsuperscript{st} 2011 to December 31\textsuperscript{st} 2011 for validation; and January 1\textsuperscript{st} 2012 to December 31\textsuperscript{st} 2012 for testing.

The off-line predictors for both SAFE and time-domain FE in each dataset are identical. For the IBM dataset, the DNN consists of 3 hidden layers, where each layer has 100 neurons. To avoid over-fitting, we implemented drop-out regularizers with rate equals to 0.1. The training scheme of the DNN is similar to that of explained in the previous section. Furthermore, before the online predictor is applied, the parameters of the SAFE approach and time-domain FE have to be set. The warning and trigger thresholds for the SAFE approach are 0.025 and 0.5 respectively while the warning and trigger thresholds for the time-domain FE are 0.27 and 0.55, respectively. The proportional deviation gain for both the approaches is set to 0.1.

The results of the proposed approach on IBM dataset is summarized in Table~\ref{tab:1b}. In general, the results show that higher percentage update leads to significantly better MSE, but the execution time also increases significantly. The table also shows that SAFE produces errors around three times lower than time-domain FE does, and executes the experiments faster than time-domain FE does. However, both of the approaches are able to provide acceptable performances compare to the baseline predictor.

\begin{figure}[!htbp]
\centering
\includegraphics[width=0.5\textwidth]{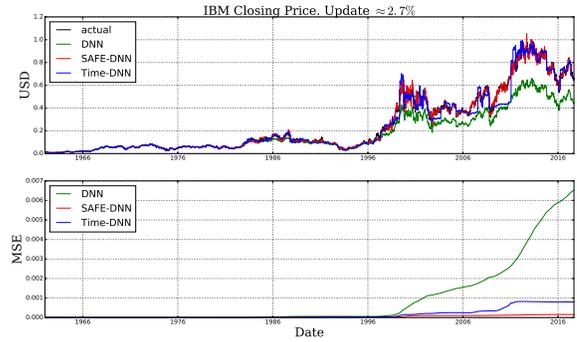}
\caption{IBM dataset performances with percentage update $\approx 2.7$\%}
\label{fig:IBM_27}
\end{figure}

Figure~\ref{fig:IBM_27} illustrates the performance comparisons between the baseline predictor, SAFE-DNN, and time-domain FE-DNN in terms of the time-series prediction and MSE. We present a case where the percentage update is around 2.7\%. We can see from the figure that if the predictor does not adapt to the non-stationarity, the time-series prediction drifts away from the actual values. This is not the case on both the approaches. In term of the errors, our proposed approach maintains low error throughout the experiments. The competing approach, however, drifts a little bit at the end when the prediction becomes more difficult due to the highly non-stationarity in the data.

\begin{figure}[!htbp]
\centering
\includegraphics[width=0.5\textwidth]{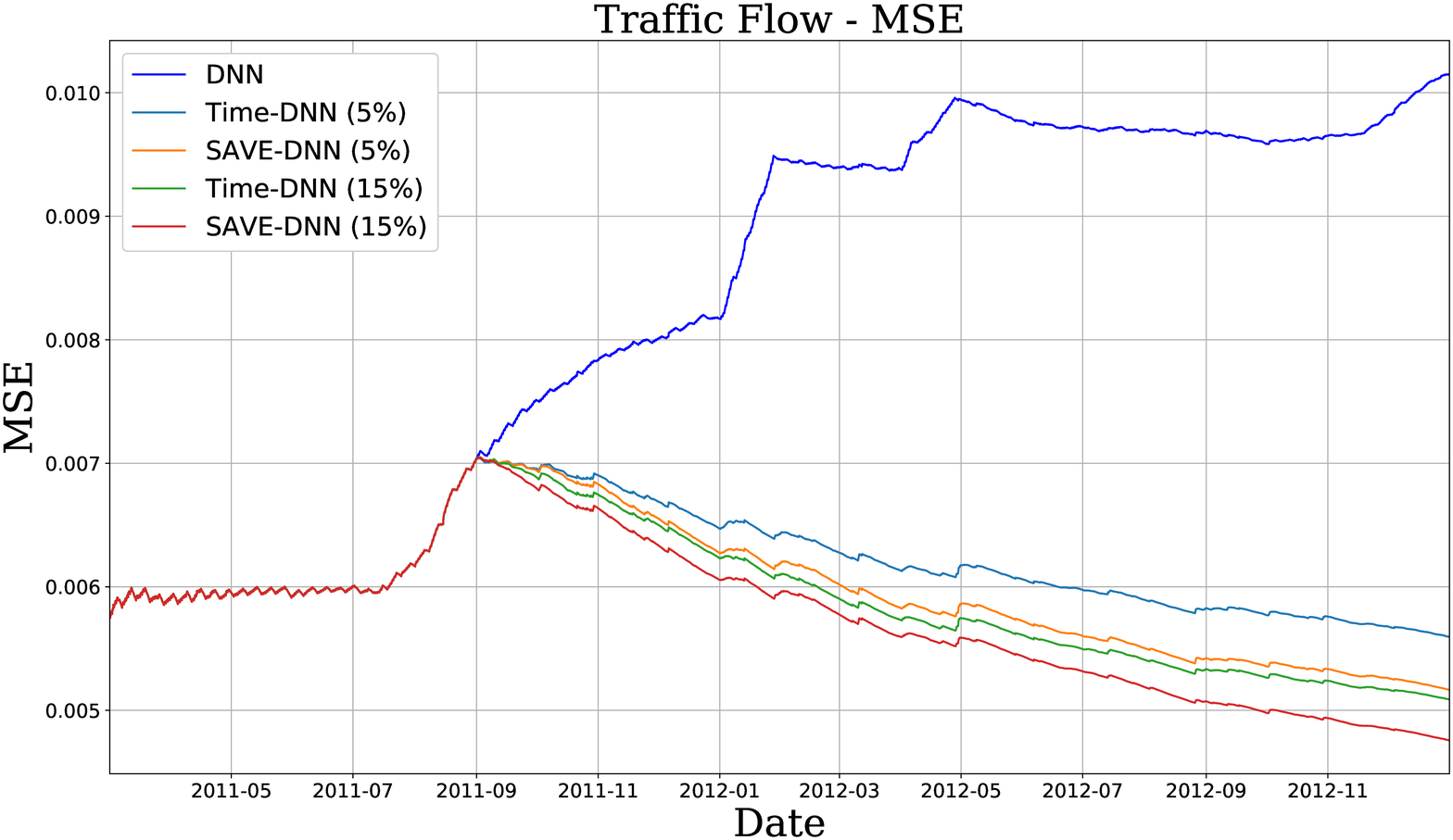}
\caption{The evolution of errors on TF dataset.}
\label{fig:TF_mse}
\end{figure}

Next, we test our approach with traffic flow dataset. The baseline DNN is configured as follows: the number of hidden layers is 3; each hidden layer contains 125 neurons; and the drop-out rate equals to 0.5. The activation function of both the hidden layers and the output layer is ReLU since we know that the traffic flow values cannot be negative. We select 5 freeways in our experiments. The results are shown in Table~\ref{tab:1b}. The results illustrate similar behavior as the ones on IBM datasets. In general, the combination of SAFE and DNN produces superior results in terms of errors and execution time.The evolution of the errors is shown in Figure~\ref{fig:TF_mse}. It can be seen that the errors start at the same level. However, the baseline error drifts away while the online predictor's error keeps decreasing, and the SAFE-DNN error is always the lowest compared to the other errors.

\begin{figure}[!htbp]
\centering
\includegraphics[width=0.5\textwidth]{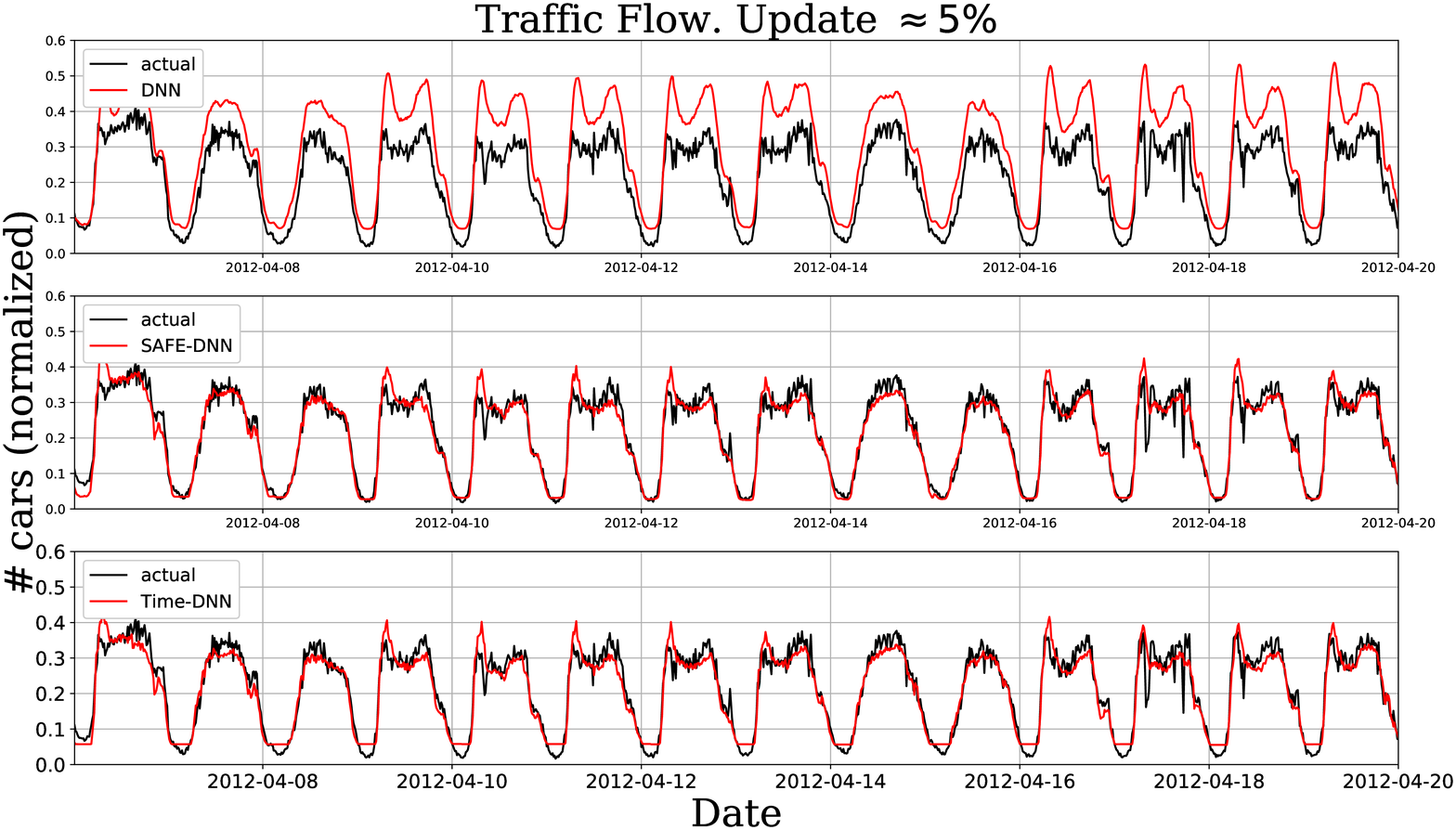}
\caption{Traffic flow prediction with percentage of update $\approx$ 5\%.}
\label{fig:TF_pred}
\end{figure}

Lastly, Figure~\ref{fig:TF_pred} shows the prediction of traffic flow time-series using baseline predictor, SAFE approach, and time-domain FE approach. We select a portion of the prediction to better illustrate the results. It can be seen that the baseline predictor does not produce acceptable traffic flow prediction while the SAFE-DNN and time-domain FE-DNN do. Although the predictions of both the approaches look similar, they are essentially different, especially in the valley parts of the traffic flow. It should be noted that this excellent prediction is obtained only by updating the predictor 5\% throughout the experiments. This means we save around 95\% of the processor or GPU cycles.

\section{Conclusion\label{conclusion}}
This paper presents an approach to actively detect non-stationarity for time series prediction. The approach monitors the evolution of the spectral contents of time series using a distance function. We have successfully conducted comprehensive experiments to validate our hypothesis and test the effectiveness of our proposed approach on artificial and real-world datasets.

The experiments show that the approach is able to achieve high long-term prediction performances while significantly saving computational resources in terms of processor and GPU cycles. Although DNN requires more computational time than other predictors do, it is clearly worth to consider as an online predictor since its overall prediction errors are notably lower than those of the other predictors. The implementation of the proportional algorithm to variably include some data in the past makes the online adaptation of the predictors more flexible, i.e., there is no need to fix the batch size of the online training procedure.

To go further with this research, we can expand the approach to work with multi-step time series predictions. Furthermore, since Long-Short Term Memory recurrent neural networks are powerful in handling sequential data, this type of neural networks is worth to investigate. Moreover, the proposed method can be used to work with large-scale time-series, where distributed neural networks, i.e., DNN with multitask learning, are appropriate.
\bibliography{reference-nonstationary}

\begin{thebibliography}{10}

\bibitem{timeseries2017}
DB-engines, {\em DBMS popularity broken down by database model}, 2017 (accessed
  October 8, 2017).

\bibitem{nason2006stationary}
G.~P. Nason, ``Stationary and non-stationary time series,'' {\em Statistics in
  Volcanology. Special Publications of IAVCEI}, vol.~1, pp.~000--000, 2006.

\bibitem{priestley1969test}
M.~Priestley and T.~S. Rao, ``A test for non-stationarity of time-series,''
  {\em Journal of the Royal Statistical Society. Series B (Methodological)},
  pp.~140--149, 1969.

\bibitem{adak1998time}
S.~Adak, ``Time-dependent spectral analysis of nonstationary time series,''
  {\em Journal of the American Statistical Association}, vol.~93, no.~444,
  pp.~1488--1501, 1998.

\bibitem{andreou2009structural}
E.~Andreou and E.~Ghysels, ``Structural breaks in financial time series,'' {\em
  Handbook of financial time series}, pp.~839--870, 2009.

\bibitem{berkes2009testing}
I.~Berkes, E.~Gombay, and L.~Horv{\'a}th, ``Testing for changes in the
  covariance structure of linear processes,'' {\em Journal of Statistical
  Planning and Inference}, vol.~139, no.~6, pp.~2044--2063, 2009.

\bibitem{cho2012multiscale}
H.~Cho and P.~Fryzlewicz, ``Multiscale and multilevel technique for consistent
  segmentation of nonstationary time series,'' {\em Statistica Sinica},
  pp.~207--229, 2012.

\bibitem{korkas2017multiple}
K.~K. Korkas and P.~Fryzlewicz, ``Multiple change-point detection for
  non-stationary time series using wild binary segmentation,'' {\em Statistica
  Sinica}, vol.~27, no.~1, pp.~287--311, 2017.

\bibitem{dwivedi2011test}
Y.~Dwivedi and S.~Subba~Rao, ``A test for second-order stationarity of a time
  series based on the discrete fourier transform,'' {\em Journal of Time Series
  Analysis}, vol.~32, no.~1, pp.~68--91, 2011.

\bibitem{fdez2007applying}
F.~Fdez-Riverola, E.~L. Iglesias, F.~D{\'\i}az, J.~R. M{\'e}ndez, and J.~M.
  Corchado, ``Applying lazy learning algorithms to tackle concept drift in spam
  filtering,'' {\em Expert Systems with Applications}, vol.~33, no.~1,
  pp.~36--48, 2007.

\bibitem{ross2012exponentially}
G.~J. Ross, N.~M. Adams, D.~K. Tasoulis, and D.~J. Hand, ``Exponentially
  weighted moving average charts for detecting concept drift,'' {\em Pattern
  Recognition Letters}, vol.~33, no.~2, pp.~191--198, 2012.

\bibitem{gonccalves2013rcd}
P.~M. Gon{\c{c}}alves~Jr and R.~S.~M. De~Barros, ``Rcd: A recurring concept
  drift framework,'' {\em Pattern Recognition Letters}, vol.~34, no.~9,
  pp.~1018--1025, 2013.

\bibitem{ditzler2015learning}
G.~Ditzler, M.~Roveri, C.~Alippi, and R.~Polikar, ``Learning in nonstationary
  environments: A survey,'' {\em IEEE Computational Intelligence Magazine},
  vol.~10, no.~4, pp.~12--25, 2015.

\bibitem{kolter2007dynamic}
J.~Z. Kolter and M.~A. Maloof, ``Dynamic weighted majority: An ensemble method
  for drifting concepts,'' {\em Journal of Machine Learning Research}, vol.~8,
  no.~Dec, pp.~2755--2790, 2007.

\bibitem{guajardo2010model}
J.~A. Guajardo, R.~Weber, and J.~Miranda, ``A model updating strategy for
  predicting time series with seasonal patterns,'' {\em Applied Soft
  Computing}, vol.~10, no.~1, pp.~276--283, 2010.

\bibitem{elwell2011incremental}
R.~Elwell and R.~Polikar, ``Incremental learning of concept drift in
  nonstationary environments,'' {\em IEEE Transactions on Neural Networks},
  vol.~22, no.~10, pp.~1517--1531, 2011.

\bibitem{moreira2016concept}
L.~Moreira-Matias, J.~Gama, and J.~Mendes-Moreira, ``Concept neurons--handling
  drift issues for real-time industrial data mining,'' in {\em Joint European
  Conference on Machine Learning and Knowledge Discovery in Databases},
  pp.~96--111, Springer, 2016.

\bibitem{cavalcante2016fedd}
R.~C. Cavalcante, L.~L. Minku, and A.~L. Oliveira, ``Fedd: Feature extraction
  for explicit concept drift detection in time series,'' in {\em Neural
  Networks (IJCNN), 2016 International Joint Conference on}, pp.~740--747,
  IEEE, 2016.

\bibitem{alippi2011just}
C.~Alippi, G.~Boracchi, and M.~Roveri, ``A just-in-time adaptive classification
  system based on the intersection of confidence intervals rule,'' {\em Neural
  Networks}, vol.~24, no.~8, pp.~791--800, 2011.

\bibitem{liu2013change}
S.~Liu, M.~Yamada, N.~Collier, and M.~Sugiyama, ``Change-point detection in
  time-series data by relative density-ratio estimation,'' {\em Neural
  Networks}, vol.~43, pp.~72--83, 2013.

\bibitem{alippi2013just}
C.~Alippi, G.~Boracchi, and M.~Roveri, ``Just-in-time classifiers for recurrent
  concepts,'' {\em IEEE transactions on neural networks and learning systems},
  vol.~24, no.~4, pp.~620--634, 2013.

\bibitem{french1999catastrophic}
R.~M. French, ``Catastrophic forgetting in connectionist networks,'' {\em
  Trends in cognitive sciences}, vol.~3, no.~4, pp.~128--135, 1999.

\bibitem{rusu2016progressive}
A.~A. Rusu, N.~C. Rabinowitz, G.~Desjardins, H.~Soyer, J.~Kirkpatrick,
  K.~Kavukcuoglu, R.~Pascanu, and R.~Hadsell, ``Progressive neural networks,''
  {\em arXiv preprint arXiv:1606.04671}, 2016.

\bibitem{polikar2001learn++}
R.~Polikar, L.~Upda, S.~S. Upda, and V.~Honavar, ``Learn++: An incremental
  learning algorithm for supervised neural networks,'' {\em IEEE transactions
  on systems, man, and cybernetics, part C (applications and reviews)},
  vol.~31, no.~4, pp.~497--508, 2001.

\bibitem{fernando2017pathnet}
C.~Fernando, D.~Banarse, C.~Blundell, Y.~Zwols, D.~Ha, A.~A. Rusu, A.~Pritzel,
  and D.~Wierstra, ``Pathnet: Evolution channels gradient descent in super
  neural networks,'' {\em arXiv preprint arXiv:1701.08734}, 2017.

\bibitem{hashimoto2016joint}
K.~Hashimoto, C.~Xiong, Y.~Tsuruoka, and R.~Socher, ``A joint many-task model:
  Growing a neural network for multiple nlp tasks,'' {\em arXiv preprint
  arXiv:1611.01587}, 2016.

\bibitem{kirkpatrick2017overcoming}
J.~Kirkpatrick, R.~Pascanu, N.~Rabinowitz, J.~Veness, G.~Desjardins, A.~A.
  Rusu, K.~Milan, J.~Quan, T.~Ramalho, A.~Grabska-Barwinska, {\em et~al.},
  ``Overcoming catastrophic forgetting in neural networks,'' {\em Proceedings
  of the National Academy of Sciences}, p.~201611835, 2017.

\bibitem{crammer2006online}
K.~Crammer, O.~Dekel, J.~Keshet, S.~Shalev-Shwartz, and Y.~Singer, ``Online
  passive-aggressive algorithms,'' {\em Journal of Machine Learning Research},
  vol.~7, no.~Mar, pp.~551--585, 2006.

\bibitem{bousquet2008tradeoffs}
O.~Bousquet and L.~Bottou, ``The tradeoffs of large scale learning,'' in {\em
  Advances in neural information processing systems}, pp.~161--168, 2008.

\bibitem{PeMS}
C.~D. of~Transportation, ``{Caltrans Performance Measurement System}.''
  \url{http://pems.dot.ca.gov/}, 2016.
\newblock "[Online; accessed June-2016]".

\bibitem{manual2010volumes}
H.~C. Manual, ``Volumes 1-4.(2010),'' {\em Transporation Research Board}, 2010.

\bibitem{rahimi2008random}
A.~Rahimi and B.~Recht, ``Random features for large-scale kernel machines,'' in
  {\em Advances in neural information processing systems}, pp.~1177--1184,
  2008.

\end{thebibliography}
\bibliographystyle{ieeetr}
\end{document}